\documentclass[10pt,twocolumn,letterpaper]{article}

\usepackage[pagenumbers]{wacv} %

\definecolor{wacvblue}{rgb}{0.21,0.49,0.74}
\usepackage[pagebackref,breaklinks,colorlinks,allcolors=wacvblue]{hyperref}

\def\confName{WACV}
\def\confYear{2026}

\title{ControlEvents: Controllable Synthesis of Event Camera Data \\ with Foundational Prior from Image Diffusion Models}
\author{
    Yixuan Hu\textsuperscript{1,*} \qquad
    Yuxuan Xue\textsuperscript{2,3,*,$\dagger$} \qquad
    Simon Klenk\textsuperscript{1,5} \qquad \\
    Daniel Cremers\textsuperscript{1,5} \qquad
    Gerard Pons-Moll\textsuperscript{2,3,4}
\\[0.4ex]
{\small\textsuperscript{1~}Technical University of Munich\quad
\textsuperscript{2~}University of Tübingen\quad
	 \textsuperscript{3~}Tübingen AI Center\quad } \\
{\small \textsuperscript{4~}Max Planck Institute for Informatics, Saarland Informatics Campus\quad}
         \\ 
{\small \textsuperscript{5~} Munich Center for Machine Learning (MCML)}
{\small\quad \textsuperscript{*~}co-first author  \quad \textsuperscript{$\dagger$~}corresponding author} \\
{\small\href{https://yuxuan-xue.com/controlevents}{https://yuxuan-xue.com/controlevents}}\\
}

\makeatletter
\let\@oldmaketitle\@maketitle%
\renewcommand{\@maketitle}{
	\@oldmaketitle%
	\begin{center}
 	\includegraphics[width=\linewidth]{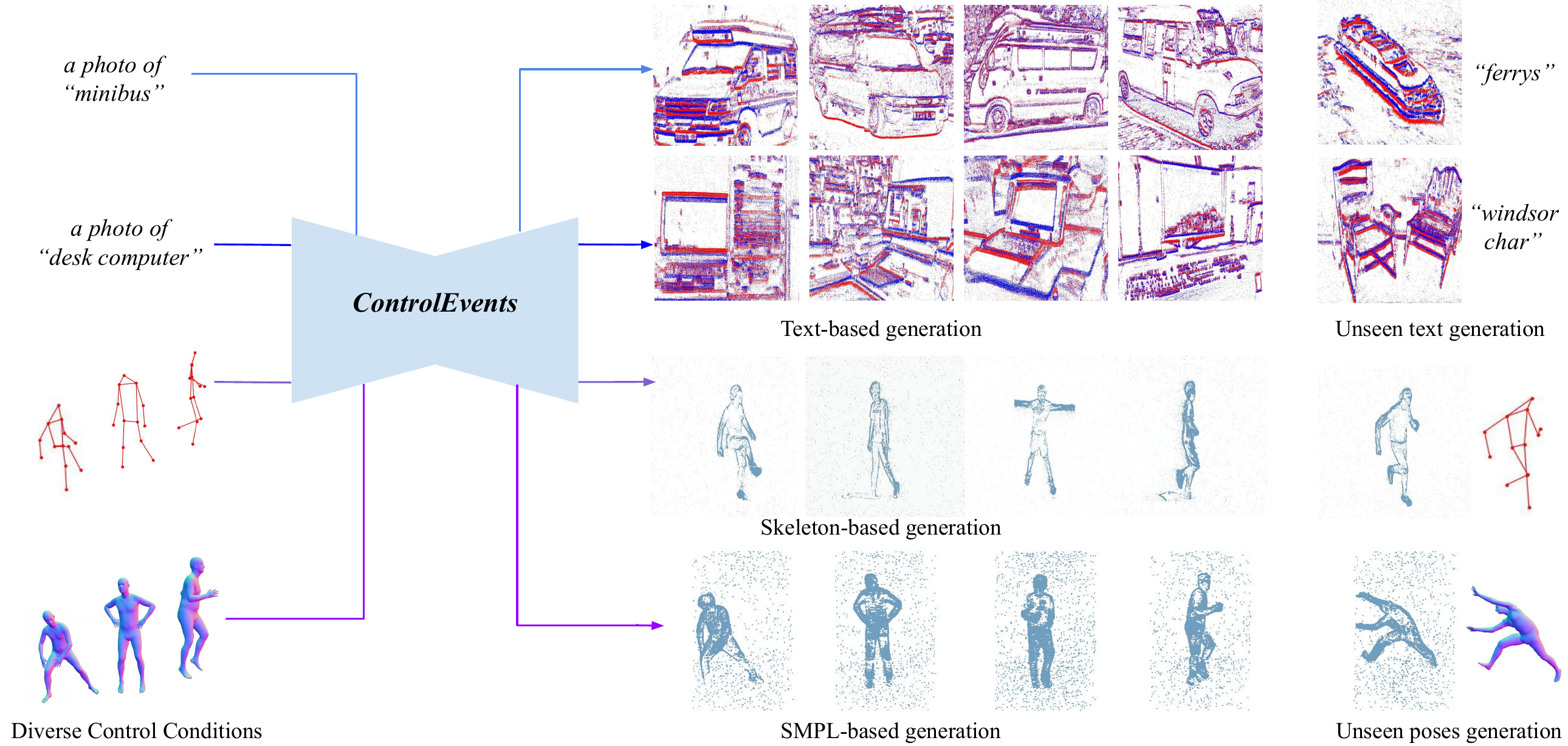}
	\end{center}
 \vspace{-3mm}
    \refstepcounter{figure}\normalfont \footnotesize
Figure~\thefigure. \textbf{\textit{ControlEvents}} can synthesize realistic event camera data from diverse conditions, such as class text label, 2D human skeleton, 3D body poses. 
\textit{ControlEvents} can generate large-scale realistic event data with pseudo ground-truth labels, which can enhance the deep learning model performance. 
	\label{fig:teaser}
\vspace{5mm}
}

\begin{document}
\maketitle
\begin{abstract}
In recent years, event cameras have gained significant attention due to their bio-inspired properties, such as high temporal resolution and high dynamic range.
However, obtaining large-scale labeled ground-truth data for event-based vision tasks remains challenging and costly, which hinders the development of the event-based algorithm.
In this paper, we present \textbf{\textit{ControlEvents}}, a diffusion-based generative model designed to synthesize unlimited high-quality event data guided by diverse control signals such as class text labels, 2D skeletons, and 3D body poses.
Our key insight is to leverage the diffusion prior from foundation models, such as Stable Diffusion, enabling high-quality event data generation with minimal fine-tuning and limited labeled data.
Our method streamlines the data generation process and significantly reduces the cost of producing labeled event datasets. 
We demonstrate the effectiveness of our approach by synthesizing event data for visual recognition, 2D skeleton estimation, and 3D body pose estimation.  
Our experiments show that the synthesized labeled event data enhances model performance in all tasks.
Additionally, our approach can generate events based on unseen text labels during training, illustrating the powerful text-based generation capabilities inherited from foundation models. 
Our models and generated datasets is publicly available for future research.

\end{abstract}
    
\section{Introduction}
\label{sec:intro}
Visual recognition and body pose estimation in challenging scenarios, such as fast-motion and low-light conditions, are critical tasks in the field of computer vision. 
These applications are relevant in diverse fields, including autonomous driving, robotics, and human-computer interaction. 
However, conventional frame-based cameras often struggle in these environments due to limitations in temporal resolution and dynamic range, resulting in motion blur and reduced image quality which hinder performance. 

In recent years, event cameras~\cite{lichtsteiner2008eventcameras} have emerged as a promising alternative to frame-based cameras.
Unlike traditional cameras, event cameras capture pixel-level changes in intensity rather than static frames, which enables them to operate effectively in high-speed and high-dynamic-range scenes~\cite{gallego_event_survey2018}. 
These features make event cameras ideal for capturing visual data in challenging conditions.

\begin{figure}[t]
    \centering
    \captionsetup{type=figure}
    \includegraphics[width=\linewidth]{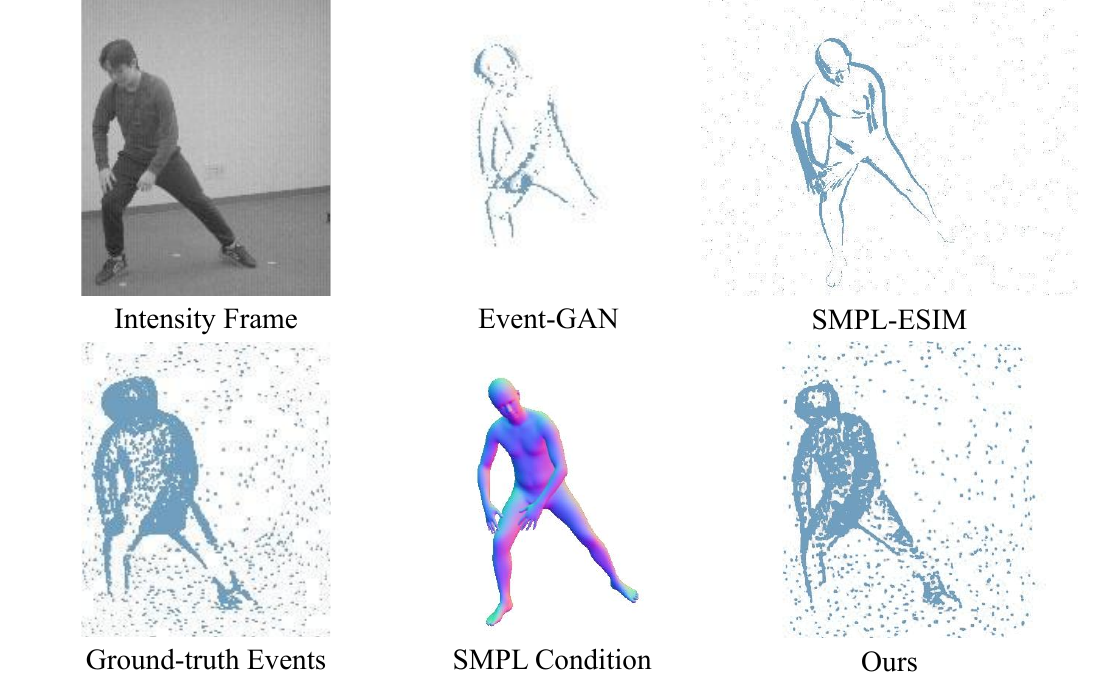}
    \vspace{-7mm}
    \caption{ \footnotesize Comparison with Evant-GAN~\cite{zhu2021eventgan} and SMPL-ESIM~\cite{xue2024elnr} on 3D human-based events generation tasks. Our \textbf{\textit{ControlEvents}} can effectively synthesize both events and noise, minimizing domain gap and closely matching the characteristics of real event data.}
    \label{fig:intro}
    \vspace{-6mm}
\end{figure}

Despite the advantages of event cameras, progress toward large-scale deep learning for event-based vision remains constrained by the scarcity of labeled ground-truth. Data annotation is time-intensive and costly, impeding robust model development. Existing simulators, such as ESIM~\cite{Rebecq18esim} for rigid motion; EventHands~\cite{rudnev2021eventhands} and SMPL-ESIM~\cite{xue2022events, xue2024elnr} for human movement,  help but fail to capture key sensor idiosyncrasies—such as background activity noise and spatially varying brightness thresholds—leading to a pronounced domain gap between simulated and real event streams (see~\cref{fig:intro}). They also suffer from slow simulation, heavy reliance on curated 3D assets, and limited labeling flexibility. Moreover, most pipelines are constrained to paired RGB images or videos as inputs and cannot generate events under diverse conditioning modalities (e.g., class, text labels), which further restricts their utility for synthesizing labeled data for event-based visual recognition.

To address this gap, we introduce \textit{ControlEvents}, a diffusion-based framework which can synthesize high-quality event data guided by diverse control signals. {\textit{ControlEvents}} can generate labeled event data conditioned on various inputs, such as class text labels~\cite{kim2021nimagenet, orchard2015ncaltech}, 2D skeletons~\cite{calabrese2019dhp19, shao2024cdehp}, and 3D body poses~\cite{zou2021eventhpe}. 
To address the challenge of limited labeled event data, {\textit{ControlEvents}} leverages the diffusion prior and generation capabilities of visual foundation models, such as Stable Diffusion~\cite{rombach2021stablediffusion} or ControlNet~\cite{zhang2023controlnet}. 
By fine-tuning pre-trained foundation models, our method can synthesize realistic, high-quality event data with minimal training and limited labeled data.
By learning directly on real captured events, our model can generate realistic noisy event data, making it more applicable for training machine learning models that generalize well to real-world data. 
Compared to traditional event simulators~\cite{Rebecq18esim, rudnev2021eventhands, xue2022events, xue2024elnr}, {\textit{ControlEvents}} simplifies data generation and significantly reduces the costs of creating large-scale, realistic, and labeled event datasets.
In summary, our contributions are as follows:
\begin{itemize}
    \item We propose  {\textit{ControlEvents}}, a diffusion-based generative model that synthesizes realistic event camera data guided by diverse control signals, which can address the scarcity of labeled data for event-based vision tasks.
    \item We demonstrate the effectiveness of {\textit{ControlEvents}} by synthesizing labeled event data for downstream tasks, achieving improved performance in visual recognition, 2D skeleton estimation, and 3D body pose estimation.
    \item We show {\textit{ControlEvents}} can generate event data based on unseen text labels during training, showcasing its text-based generation capabilities and potential for zero-shot visual recognition tasks.
    \item We make both our model and large-scale synthesized datasets publicly available, which could foster research in event-based vision tasks.
\end{itemize}

\section{Related Works}
\label{sec:related}

\paragraph{Event-based Vision} 
Unlike frame-based cameras, which sample a scene at fixed intervals as a sequence of intensity frames, event cameras operate independently at the pixel level, detecting changes in brightness and generating event streams rather than conventional images.
Event cameras capture asynchronous pixel-wise brightness changes, mimicking how the retina processes visual information~\cite{lichtsteiner2008eventcameras,gallego_event_survey2018}. 
Due to high temporal resolution, low latency, high dynamic range, and efficient power usage, event cameras are particularly suited to dynamic applications. 
However, their distinct outputs challenge traditional computer vision algorithms, which are designed for frame-based data. 
Recent advancements have adapted event-based cameras to various computer vision tasks~\cite{gallego_event_survey2018}. 
Applications in optical flow estimation~\cite{bardow2016event_optical_flow, Gehrig2021eraft, zhu2018unsupervised_optical_flow}, SLAM~\cite{kim2016_ev3Drecon, rebecq2017_evvio, vidal2018_ultimateslam, bryner2019event_generation_model}, and 3D body recovery~\cite{xue2022events, xue2024elnr, rudnev2021eventhands}, demonstrate the potential of event cameras.

Despite above advantages, labeling ground truth for event data is challenging, which hinders the training of deep learning models for event-based vision tasks. 
To address this, event simulators~\cite{Rebecq18esim, rudnev2021eventhands, xue2024elnr, xue2022events} generate synthetic events by subtracting consecutive frames from RGB renderings. 
Unfortunately, these simulators often results in a significant domain gap from real-world events. 
In this work, we leverage pre-trained foundation models~\cite{zhang2023controlnet, rombach2021stablediffusion} to synthesize realistic event data, enabling producing infinite labeled events data from a limited real-world labeled events.

\vspace{-0.4cm}
\paragraph{Synthetic Dataset Generation} 
Synthetic datasets are increasingly used to address tasks with limited ground-truth data. 
This approach is prominent in optical flow estimation~\cite{varol17surreal}, point tracking~\cite{zheng2023pointodyssey}, depth estimation~\cite{savva2019habitat, roberts2021hypersim}, and 3D body estimation~\cite{varol17surreal, black2023bedlam, yang2023synbody, cai2024gtahuman}. 
Most synthetic datasets are generated via computer graphics engines. 
For example, Infinigen~\cite{infinigen2023infinite, infinigen2024indoors} uses Blender to render photorealistic scenes, and BEDLAM~\cite{black2023bedlam} leverages the Unreal Engine to create human movements with realistic clothing dynamics, providing accurate ground-truth for body pose and shape in SMPL~\cite{loper2015smpl, pavlakos2019smplx} format. 
These works underscore the potential of graphics engines for generating detailed synthetic data. However, high-quality 3D assets are required, which can be costly and time-intensive.

Recent progress in deep generative models has facilitated image synthesis directly from control signals, bypassing the need for extensive 3D assets. 
Methods using pre-trained models like Stable Diffusion~\cite{rombach2021stablediffusion} have enabled the generation of controlled images with minimal fine-tuning for depth and normal maps using ControlNet~\cite{zhang2023controlnet}. STAGE~\cite{kister2024stage} synthesizes human images from skeletons and depth cues, while 3D-DST~\cite{ma20243ddst} uses 3D CAD models for controlled generation. 
These studies illustrate that generative models can produce realistic images under explicit control, and generated data could further enhancing the performance of machine learning models. 
However, no existing work explores controllable generation for events data, which exhibits a considerable domain gap from RGB images.

For event data generation, simulators like ESIM~\cite{Rebecq18esim} generate events from rigid object motion, supporting applications in SLAM and object tracking. More recent simulators, such as EventHands~\cite{rudnev2021eventhands} and E-LnR~\cite{xue2024elnr, xue2022events}, focus on non-rigid deformations like human body movement. However, these simulators struggle to simulate the noise and artifacts of real-world event cameras, leading to domain discrepancies between synthetic and actual event data.
By contrast, our {\textit{ControlEvents}} leverages the diffusion prior from~\cite{rombach2021stablediffusion, zhang2023controlnet} and simulates characteristics of real-world captured events. Experiments in~\cref{fig:intro} and~\cref{sec:experiments} show that our generated events outperform previous simulators and can be used as training data to improve the visual recognition and 2D \& 3D pose estimation performance.

\section{Method}
\label{sec:method}

\begin{figure*}[t]
    \centering
    \captionsetup{type=figure}
    \includegraphics[width=\linewidth]{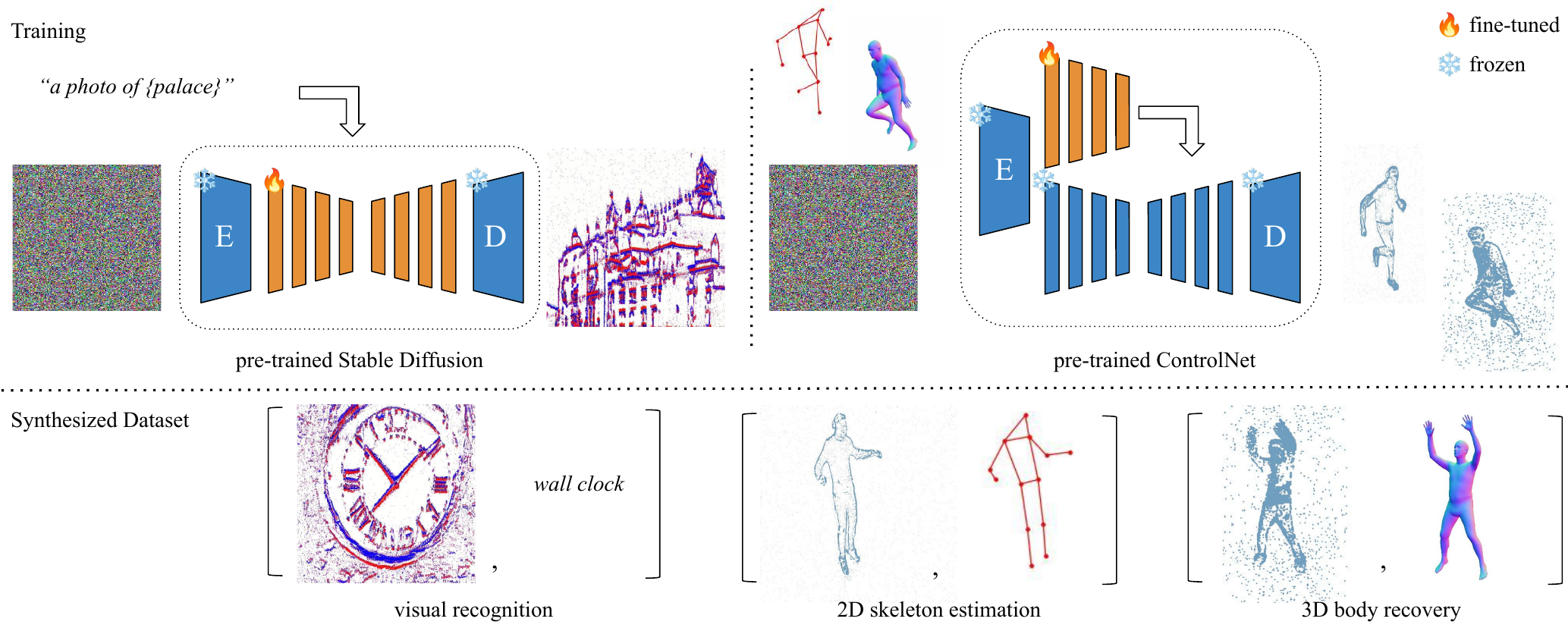}
    \vspace{-7mm}
    \caption{ \footnotesize Overview of \textbf{\textit{ControlEvents}}. For text-conditioned event data synthesis, we fine-tune Stable Diffusion~\cite{rombach2021stablediffusion}. For 2D \& 3D pose-conditioned event data synthesis, we fine-tune ControlNet~\cite{zhang2023controlnet} using skeleton map or normal map. Our \textit{ControlEvents} can synthesize large-scale dataset for various tasks.}
    \vspace{-5mm}
\end{figure*}

\subsection{Preliminaries}
\label{sec:preliminaries}

\paragraph{Event Camera model}
Event cameras record logarithmic pixel-level brightness change asynchronously.  
An event $e_i = (\mathbf{u}_i, t_i, p_i)$ is triggered at pixel $\mathbf{u}_i$ at time $t_i$ when the logarithmic brightness change $\Delta\mathcal{L}$ reaches a threshold. The polarity of event $e_i$ is  
\begin{equation}
    p(\mathbf{u}_i, t_i) = 
    \begin{cases}
      $+1$ & \text{if $\mathcal{L}\left(\mathbf{u}_i, t_{i}\right) - \mathcal{L}\left(\mathbf{u}_i, t_{i-1}\right) \geq C $ }, \\
      $-1$ & \text{if $\mathcal{L}\left(\mathbf{u}_i, t_{i-1}\right) - \mathcal{L}\left(\mathbf{u}_i, t_{i}\right) \geq C $ }, \\
    \end{cases}       
\end{equation}
where $C$ is the contrastive threshold, a hardware-specific property of event cameras~\cite{lichtsteiner2008eventcameras}.

To fully leverage the capabilities of pre-trained image diffusion models~\cite{rombach2021stablediffusion, zhang2023controlnet}, we convert asynchronous events into 2D histograms, which can be represented as a 3-channel RGB image (H×W×3) with positive events shown in red and negative events in blue. 
We accumulate events into frames based on the strategy used in the adopted datasets~\cite{kim2021nimagenet, calabrese2019dhp19, shao2024cdehp, zou2021eventhpe} for our downstream tasks.

In N-ImageNet~\cite{kim2021nimagenet}, events are generated by using an event camera to capture original ImageNet images displayed on a screen. Hence, we accumulate all events for each image. 
For DHP-19~\cite{calabrese2019dhp19}, we accumulate a fixed number (7,500) of events per frame. For CDEHP~\cite{shao2024cdehp}, events are accumulated over a fixed time interval (8.333 ms) per frame. In Event-HPE~\cite{zou2021eventhpe}, events are aggregated over a fixed temporal length (15 ms) per frame. 
Since DHP-19, CDEHP, and Event-HPE do not provide polarity information, we represent event locations in white within the 2D histogram. 
In the paper, we visualize event images with a white background and blue color for enhanced appearance and reduced printing costs. 
Please refer to~\cref{fig:teaser} and the supplementary material for visualizations of the event frames.

\vspace{-4mm}
\paragraph{Diffusion Models under Control}
DDPM~\cite{Ho2020DDPM} is a generative model which learns a data distribution by iteratively adding and removing noise. 
Formally, the forward process in DDPM iteratively adds noise to a sample $\mathbf{x}_0$ drawn from a distribution $p_\text{data}(\mathbf{x})$:
\begin{equation}
\begin{gathered}
    \mathbf{x}_t \sim \mathcal{N}(\mathbf{x}_t; \sqrt{\alpha_t} \mathbf{x}_{t-1}, (1-\alpha_t)\mathbf{I}),
\end{gathered}
\label{eq:ddpm_forward_1}
\end{equation}
and $\alpha_t$ schedules the amount of noise added at each step $t$~\cite{Ho2020DDPM}. 
To sample data from the learned distribution, the reverse process starts from Gaussian noise $\mathbf{x}_T\sim\mathcal{N}(0, \mathbf{I})$ and iteratively denoises it until $t=0$: 
\begin{equation}
\begin{gathered}
    \mathbf{x}_{t-1}\sim\mathcal{N}(\mathbf{x}_{t-1};\mathbf{\mu}_\theta(\mathbf{x}_t, t), \frac{1-\bar{\alpha}_{t-1}}{1-\bar{\alpha}_{t}} (1-\alpha_t)\mathbf{I}),
    \label{eq:ddpm_reverse_1}
\end{gathered}
\end{equation}
where $\mathbf{\mu}_\theta(\mathbf{x}_t, t)$ is the estimated posterior mean, predicted by a neural network $\epsilon_{\theta}$ trained to minimize the difference between true and estimated noise.
One can also model conditional distribution $p_\text{data}(\mathbf{x} | \mathbf{c})$ by adding the condition to the network input~\cite{dhariwal2021diffusionClassGuidance, ho2021classifierfreeCFG}. In practice, the network $\epsilon_{\theta}$ predicts the amount of noise to remove based on the noisy state $\mathbf{x}_t$, condition $x$, and diffusion timestep $t$, optimizing an objective function:
\begin{equation}
    L_{\text{objective}}=\sum_{t=1}^T \mathbb{E}_{\mathbf{x}_0, \mathbf{\epsilon}, t}\left[\left\|\mathbf{\epsilon}-\mathbf{\epsilon}_\theta\left(\mathbf{x}_t, t, \mathbf{c}\right)\right\|^2\right].
    \label{eq:objective_ddpm}
\end{equation}

\subsection{ControlEvents}
Due to the scarcity of event data its ground-truth labels, training large-scale deep learning models for event-based tasks is impossible. Our goal is to improve model performance by training on synthetic event data generated by {\textit{ControlEvents}}. This procedure can be viewed as an analysis-by-synthesis loop: given a small amount of labeled real-world event data, we train a generative model to synthesize event data with pseudo ground-truth labels. This synthetic data, paired with pseudo labels, serves as additional training data to enhance the model’s analytical capabilities.
To achieve this alignment, it is essential to match the distribution of synthetic events to the real event data. However, the limited availability of event data presents a scientific challenge: \emph{how can a generative model effectively learn a robust data distribution when only a small amount of training data from the target distribution is available?}

Inspired by recent works~\cite{ke2024marigold, ye2024stablenormal, zhang2023controlnet, xue2024human3diffusion, xue2025infinihuman, xue2025gen3diffusion, xue2025physic}, we observe that the diffusion prior of foundational models (e.g., Stable Diffusion~\cite{rombach2021stablediffusion}) contains rich information that enhances the generalization capabilities of downstream applications. Thus, we design {\textit{ControlEvents}} on top of foundational generative models, allowing us to leverage their generative power to synthesize realistic event data by minimal fine-tuning on small-scale real event datasets.

We denote the available small-scale real event datasets $\mathcal{D}_{\text{real}}=\{(\mathbf{e}^i, \mathbf{c}^i)\}^{N}_{i=1}$, where $\mathbf{e}^i$ is an event image obtained by accumulating the event stream within a fixed temporal window or a fixed number of events.
The corresponding ground-truth label, $\mathbf{c}^i$, could be a class label in natural language~\cite{kim2021nimagenet, orchard2015ncaltech}, a 2D human skeleton~\cite{calabrese2019dhp19, shao2024cdehp}, or a 3D human body~\cite{shao2024cdehp} represented in SMPL~\cite{loper2015smpl, pavlakos2019smplx}, depending on the generation task.
To model the conditional data distribution $p_\text{data}(\mathbf{e} | \mathbf{c})$, we adopt the DDPM framework described in~\cref{sec:preliminaries} and formulate the objective function as 
\begin{equation}
    L_{{\textit{ControlEvents}}}=\sum_{i=1}^N\sum_{t=1}^T \mathbb{E}_{\mathbf{e}, \mathbf{\epsilon}, t}\left[\left\|\mathbf{\epsilon}-\mathbf{\epsilon}_\theta\left(\mathbf{e}^i_t, t, \mathbf{c}^i\right)\right\|^2\right].
    \label{eq:objective_controlevents}
\end{equation}
Here, $\mathbf{e}^i_t$ is a noisy image generated from the clear event image $\mathbf{e}^i$ using DDPM scheduler as outlined in~\cref{eq:ddpm_forward_1}. The term $\mathbf{\epsilon}_\theta(\cdot)$ represents the diffusion network, parametrized by parameters $\theta$, which predicts the noise level to subtract during inference. 
At inference time, given an unseen control signal $\mathbf{c}^{\text{novel}}$, we synthesize the corresponding event image $\mathbf{e}^{\text{novel}}$ by sampling from the learned distribution $p_\text{data}(\mathbf{e} | \mathbf{c})$. Specifically, we perform the iterative denoising starting from randomly sampled Gaussian noise using~\cref{eq:ddpm_reverse_1} until a clear event image is generated.

\vspace{-0.4cm}
\paragraph{Generation from Text Label}

To leverage the foundational text-to-image generative capabilities of image diffusion models~\cite{rombach2021stablediffusion}, we fine-tune a pre-trained Stable Diffusion~\cite{rombach2021stablediffusion}. Originally, it models the text-conditioned image distribution $p_\text{data}(\mathbf{x} | \mathbf{c^{\text{text}}})$. Here, we fine-tune and adapt it to the target event distribution $p_\text{data}(\mathbf{e} | \mathbf{c^{\text{class}}})$.

In an event-based classsification dataset $\mathcal{D}^{\text{class}}_{\text{real}}=\{(\mathbf{e}^i, \mathbf{c}^{\text{class}, i})\}^{N}_{i=1}$, such as N-ImageNet~\cite{kim2021nimagenet} or N-Caltech101~\cite{orchard2015ncaltech}, each event image $\mathbf{e}^i$ is associated with a class label $\mathbf{c}^{\text{class}, i}$. We enhance the class label $\mathbf{c}^{\text{class}, i}$ by augmenting it with the phrase: \textit{"A photo of $\{\mathbf{c}^{\text{class}, i}$\}"}, making it resemble a natural language description and more aligned with the text conditioning used in Stable Diffusion~\cite{rombach2021stablediffusion}.  
The event image $\mathbf{e}^i$ is a 3-channel image, with positive events encoded in the first channel (red) and negative events in the third channel (blue). 
We utilize the frozen VAE encoder~\cite{rombach2021stablediffusion} to obtain the latent representation of $\mathbf{e}^i$ and fine-tune only the U-Net diffusion model using the objective function defined in~\cref{eq:objective_controlevents}. We denote the fine-tuned text-based event generator as ${\textit{ControlEvents}}^{\text{class}}$.

After adapting the ${\textit{ControlEvents}}^{\text{class}}$ with the event-based classification dataset $\mathcal{D}^{\text{class}}_{\text{real}}$, we generate events for each class with the adapted model conditioned on the same text prompt, forming a synthetic dataset $\mathcal{D}^{\text{class}}_{\text{synth}}$. We show in~\cref{subsec:exp_classification} that the synthetic data generated by our ${\textit{ControlEvents}}^{\text{class}}$ can improve the event-based visual recognition performance. %

\vspace{-0.4cm}
\paragraph{Generation from 2D Skeleton}
Estimating 2D human pose in skeleton from event data is a valuable application, as human motion is often too fast for frame-based cameras and can cause motion blur. As a widely-used generative framework, ControlNet~\cite{zhang2023controlnet} enables the incorporation of 2D skeleton signals into the realistic image diffusion process of Stable Diffusion~\cite{rombach2021stablediffusion}. Therefore, we use $\text{ControlNet}^{\text{skeleton}}$~\cite{zhang2023controlnet} as our starting point, as it already models the image distribution $p_\text{data}(\mathbf{x} | \mathbf{c^{\text{skeleton}}})$ effectively.

Here, we use an event-based pose estimation dataset $\mathcal{D}^{\text{skeleton}}_{\text{real}}=\{(\mathbf{e}^i, \mathbf{c}^{\text{skeleton}, i})\}^{N}_{i=1}$, such as DHP-19~\cite{calabrese2019dhp19} or CDEHP~\cite{shao2024cdehp}. Since ControlNet also supports text descriptions as additional conditioning, we apply the description \textit{"A moving human"} for all event images $\mathbf{e}^{i}$. Following the training procedure of ControlNet~\cite{zhang2023controlnet}, we fine-tune only the trainable copy of the U-Net, keeping the original Stable Diffusion U-Net and VAE frozen. 
We denote the fine-tuned skeleton-based event generator as ${\textit{ControlEvents}}^{\text{skeleton}}$. 
Using this model, we generate event images from unseen 2D skeletons during training and construct a synthetic dataset $\mathcal{D}^{\text{skeleton}}_{\text{synth}}$, which we then use to train an event-based pose estimation model. Further details about the generation process and training of the 2D pose estimator are provided in~\cref{subsec:exp_2dpose}.
 
\vspace{-0.4cm}
\paragraph{Generation from 3D Body}
Estimating the parameters of a 3D body model such as SMPL~\cite{loper2015smpl, pavlakos2019smplx} from event data is more challenging than estimating a 2D skeleton due to self-occlusion and monocular ambiguity. 
Training large-scale deep learning models is a common approach, which motivates us to synthesize high-quality event data.
However, no existing image foundation models take numerical SMPL parameters (e.g., pose and shape) as direct conditioning, which makes our fine-tuning difficult. 

To fully leverage the diffusion prior of foundation models, we propose representing the 3D SMPL model using 2D images that can serve as conditioning inputs. Specifically, we choose normal maps as the 2D conditioning images for the 3D SMPL body, as they provide surface direction information at each pixel~\cite{zhu2024champ}. This design choice allows us to utilize the capabilities of pre-trained $\text{ControlNet}^{\text{normal}}$, which generates corresponding realistic images from normal map by sampling from learned distribution $p_\text{data}(\mathbf{x} | \mathbf{c^{\text{normal}}})$.

Given a real-world dataset which contains event data and corresponding SMPL parameters $\mathcal{D}^{\text{SMPL}}_{\text{real}}=\{(\mathbf{e}^i, \mathbf{c}^{\text{SMPL}, i})\}^{N}_{i=1}$, such as Event-HPE~\cite{zou2021eventhpe}, we render each SMPL parameters $\mathbf{c}^{\text{SMPL}, i}$ into a 2D normal map $\mathbf{c}^{\text{normal}, i}$  in Blender~\cite{blender} with the camera intrinsics and extrinsics from~\cite{zou2021eventhpe}. Similar to the 2D pose task, we apply the description “A moving human” to all event images $\mathbf{e}^{i}$ and fine-tune only the trainable copy of the U-Net. We denote the SMPL normal-based event generator as ${\textit{ControlEvents}}^{\text{SMPL}}$. Given such a model, we can easily generate event images from arbitrary SMPL parameters. We demonstrate that our ${\textit{ControlEvents}}^{\text{SMPL}}$ generalizes well to diverse and challenging human poses from SMPL datasets like AMASS~\cite{mahmood2019amass} and AIST~\cite{Tsuchida2019aist}. This capability allows us to obtain large-scale challenging event data with pseudo ground-truth, which is not available in existing real-world event datasets. Please refer to~\cref{subsec:exp_3dpose} for details.

\label{sec:controlevents}

\section{Experiments}
\label{sec:experiments}

In this section, we empirically demonstrate the effectiveness of our \textit{{ControlEvents}} and synthesized event data by evaluating it on discriminative downstream tasks: object recognition in~\cref{subsec:exp_classification}, 2D skeleton estimation in~\cref{subsec:exp_2dpose}, and 3D SMPL estimation in~\cref{subsec:exp_3dpose}. 
We further validate our synthetic event data using standard generative model metrics, namely Fréchet Inception Distance (FID)~\cite{heusel2017fid} in~\cref{tab:FID_3Dsmpl}

\vspace{-0.4cm}
\paragraph{Baselines}
Event simulators can be classified into two categories: learning-based approaches, such as Event-GAN~\cite{zhu2021eventgan}, which synthesize events from two consecutive intensity frames; and graphics-based approaches, such as SMPL-ESIM~\cite{xue2022events, xue2024elnr}, which simulate events from a sequence of SMPL motions. 
Consequently, we compare our method with Event-GAN in~\cref{subsec:exp_2dpose} and~\cref{subsec:exp_3dpose}, and with SMPL-ESIM~\cref{subsec:exp_3dpose}, because these methods support similar control signals for those tasks. However, neither can synthesize event data from text-based class labels, which is a unique advantage of our proposed {\textit{ControlEvents}}.

\subsection{Event-based Object Recognition}
\label{subsec:exp_classification}

\paragraph{Implementation Details}
We evaluate our method on the NImageNet dataset~\cite{kim2021nimagenet}, which contains 1,000 classes. For each class in NImageNet, multiple (typically 2-5) similar class names are available. We use only 100 event images per class for learning $p_\text{data}(\mathbf{e} | \mathbf{c^{\text{class}}})$. In each training iteration, we randomly sample a class name and augment it as \textit{"A photo of $\{\mathbf{c}^{\text{class}, i}$\}"} to serve as the text description. Please refer to supp. mat. for training details of  ${\textit{ControlEvents}}^{\text{class}}$.

\vspace{-0.4cm}
\paragraph{Classification with Generated Events}
We infer ${\textit{ControlEvents}}^{\text{class}}$ to synthesize 650 event images for each class.
We train a classifier with a pre-trained ResNet-34~\cite{he2016resnet} and randomly initialize MLP prediction head with different setting, including original training data of our generative model and our synthesized event data. It shows that by combining our synthesized event images with original training data, we obtain better classification accuracy than only using the raw event data.

\begin{table}[h]
  \centering
\begin{tabular}{l c c}
 \hline
\multicolumn{3}{c}{N-ImageNet evaluation set~\cite{calabrese2019dhp19}} \\
 \hline
Data setting & {Acc. $\uparrow$}  & {Prec. $\uparrow$} \\
 \hline
 original training data &  $  39.33 $ & $ 40.02 $  \\
\hline
\textit{Ours}   &  $ 27.00 $   & $ 23.32 $  \\
\textit{Ours} + original training data   &  $\textbf{ 45.33}$   & $\textbf{ 46.52}$ \\
\hline
\end{tabular}
\caption{\textbf{Object recognition} with different training data. Our synthetic event data can enhance classification performance.}
\label{tab:compare_classfication}
\vspace{-3mm}
\end{table}

\begin{figure}[h]
    \centering
    \captionsetup{type=figure}
    \includegraphics[width=\linewidth]{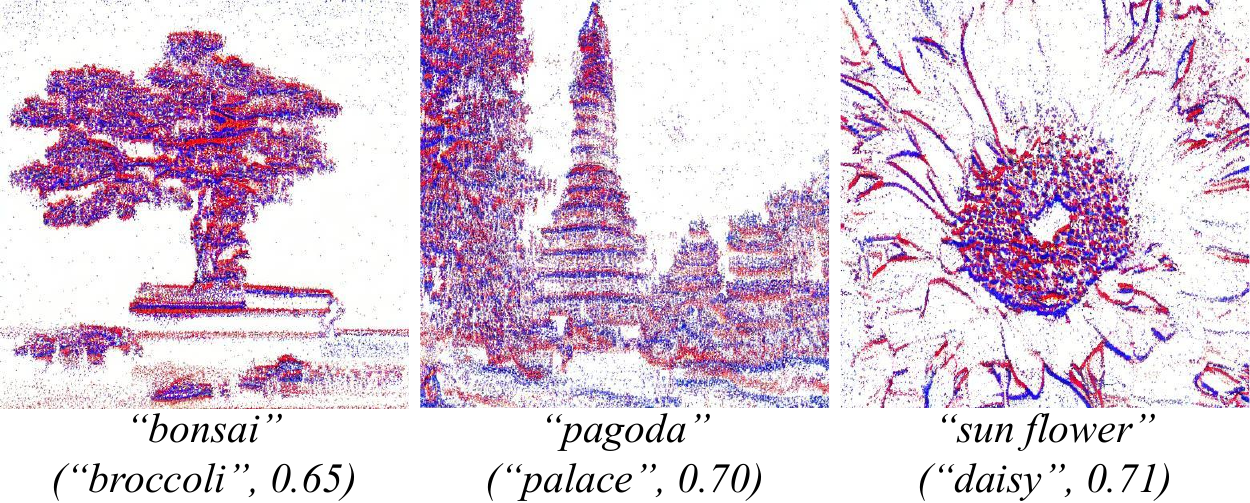}
    \vspace{-7mm}
    \caption{ \footnotesize \textbf{Zero-shot generation of unseen text label}. We also present the closest seen text label during training based on the CLIP cosine similarity.}
    \label{fig:zeroshot_our}
    
    \vspace{-5mm}
\end{figure}

\vspace{-0.4cm}
\paragraph{Zero-Shot Generation}
A key insight in {\textit{ControlEvents}} is leveraging the powerful diffusion prior in pre-trained foundation models. 
In our ${\textit{ControlEvents}}^{\text{class}}$ experiments , we utilize the robust text-based generation capabilities inherited from Stable Diffusion~\cite{rombach2021stablediffusion}. Despite training on only 1,000 classes from N-ImageNet~\cite{kim2021nimagenet}, our model generalizes to unseen text labels from the N-Caltech~\cite{orchard2015ncaltech} dataset and can generate corresponding event images. We present qualitative results of zero-shot event data generation alongside their closest training class labels (determined by cosine similarity in CLIP~\cite{radford2021clip} feature space) in~\cref{fig:zeroshot_our}. Please refer to the supplementary material for quantitative zero-shot event-based classification results on N-Caltech dataset.

\subsection{Event-based 2D Skeleton Estimation}
\label{subsec:exp_2dpose}
\paragraph{Implementation Details}
We evaluate our method on the DHP-19~\cite{calabrese2019dhp19} and CDEHP~\cite{shao2024cdehp} datasets because they provide the ground-truth 2D skeleton in the same format. We randomly sample 10000 event images with skeleton across different subjects and motion to train ${\textit{ControlEvents}}^{\text{skeleton}}$.
Please refer to supp. mat. for training details.

\vspace{-0.4cm}
\paragraph{Pose Estimation with Generated Events}
We adopt the pose estimation network architecture from~\cite{calabrese2019dhp19} and randomly initialize the network parameters during training. 
For our ${\textit{ControlEvents}}^{\text{skeleton}}$, we randomly sample 5000 unseen skeletons from DHP-19~\cite{calabrese2019dhp19} and 5000 from CDEHP~\cite{shao2024cdehp}.
Since Event-GAN~\cite{zhu2021eventgan} requires intensity frames and DHP-19~\cite{calabrese2019dhp19} does not provide them, we sample 10000 frames from CDEHP~\cite{shao2024cdehp} for generating events using Event-GAN.

We report the quantitative evaluation of 2 unseen subjects using mean per joint position error (MPJPE) and average precision (AP@25\%, AP@50\%) in~\cref{tab:compare_2d_skeleton}. The results indicate that our synthetic event data is more effective for training 2D pose estimators and can further enhance performance when combined with the original training data of ${\textit{ControlEvents}}^{\text{skeleton}}$.

\begin{table*}[t!]
  \centering
\begin{tabular}{l c c c | c c c}
 \hline
& \multicolumn{3}{c}{DHP-19 evaluation set~\cite{calabrese2019dhp19}} & \multicolumn{3}{c}{CDEHP evaluation set~\cite{shao2024cdehp}}\\
 \hline
Data Setting & {2D-MPJPE $\downarrow$}  & {AP@25 $\uparrow$} & { AP@50 $\uparrow$} &  {2D-MPJPE $\downarrow$}  & {AP@25 $\uparrow$} & { AP@50 $\uparrow$} \\
 \hline
original training data &  $ 7.180 $   & $99.5$ & $98.0$   &$12.149$ & $ 98.2 $ & $ 95.9 $  \\
Event-GAN~\cite{zhu2021eventgan}  &  $7.986$   & $98.4$ & $98.0 $ & $16.264$ &  $ 97.5 $ & $94.1$ \\
\hline
\textit{Ours}   &  $6.223$   & $\textbf{99.7}$ & $99.1$ & $10.936$ & $98.8$ & $96.5$  \\
\textit{Ours} + original training data   &  $\textbf{6.006}$   & $\textbf{99.7}$ & $\textbf{99.4}$ & $\textbf{10.792}$ & \textbf{98.9} & $\textbf{96.7}$  \\
\hline
\end{tabular}
\caption{\textbf{2D pose estimation} with different training data. Our synthetic event data outperforms~\cite{zhu2021eventgan} and enhances estimation performance.}
\label{tab:compare_2d_skeleton}
\end{table*}

\begin{table*}[t!]
  \centering
\begin{tabular}{l c c c c c}
 \hline
& \multicolumn{5}{c}{Event-HPE evaluation set~\cite{calabrese2019dhp19}} \\
 \hline
Data Setting & {3D-MPJPE$_{\text{mm}}$ $\downarrow$}  &  {3D-PCK@25 $\uparrow$}  & {3D-PCK@50 $\uparrow$}& {PVE$_{\text{mm}}$  $\downarrow$} & {AUC $\uparrow$} \\
 \hline
 original training data &  $ 27.30 $     & $ 46.05 $ & $ 95.02 $& $ 34.24$ & $ 81.79 $  \\
Event-GAN~\cite{zhu2021eventgan}  &  $29.43$  & $ 35.61$ &  $ 91.79 $ & $ 35.27$  & $ 80.37$ \\
SMPL-ESIM~\cite{xue2024elnr}  &  $53.53$    & $ 5.84 $ &  $ 42.54 $  & $ 55.35$ & $36.44$ \\
\hline
\textit{Ours w/ Event-HPE pose}   &  $25.13$   & $ 54.97 $ & $ 96.45 $ & $ 34.22$ & $ 83.24 $  \\
\textit{Ours w/ AMASS pose}   &  $28.32$   & $ 45.11 $ & $ 92.59 $ & $ 34.36$ & $ 81.12$  \\
\textit{Ours} both + original training data   &  $\textbf{18.65}$  & $\textbf{79.37}$ & $\textbf{99.66}$   & $ \textbf{32.12}$ & $ \textbf{87.57} $  \\
\hline
\end{tabular}
\caption{\textbf{3D SMPL Body Recovery} with different training data. Our synthetic event data outperforms both learning-based synthetic data~\cite{zhu2021eventgan} and graphics-based synthetic data~\cite{xue2022events, xue2024elnr}. Additionally, we demonstrate the capability to generate synthetic events for challenging poses from AMASS~\cite{mahmood2019amass}, which significantly enhances SMPL estimation performance.}
\label{tab:compare_smpl_recovery}
\vspace{-5mm}
\end{table*}

\subsection{Event-based 3D Body Recovery}
\label{subsec:exp_3dpose}
\paragraph{Implementation Details}
We evaluate our method and data on the Event-HPE~\cite{zou2021eventhpe} dataset. We randomly sample 10000 event images with SMPL annotations to train ${\textit{ControlEvents}}^{\text{SMPL}}$. Please refer to supp. mat. for details.

\vspace{-0.4cm}
\paragraph{SMPL Recovery from Generated Events}
In this experiment, we use the randomly initialized event-based SMPL regressor from~\cite{zou2021eventhpe} as the SMPL recovery model. 
For our ${\textit{ControlEvents}}^{\text{SMPL}}$, we randomly sample 10,000 unseen SMPL annotations from the Event-HPE dataset~\cite{zou2021eventhpe}. We compare our generated event images with those from Event-GAN~\cite{zhu2021eventgan} and the SMPL-ESIM simulator~\cite{xue2024elnr}. 
To minimize variations due to pose differences during generation, we use the same sampled SMPL annotations to synthesize events for both Event-GAN and SMPL-ESIM. 

We report quantitative evaluation of 2 unseen subjects on MPJPE, mean per vertex error (PVE), percent of correct keypoints (PCK@25mm, PCK@50mm), and area under PCK curve (AUC). We compare 3D SMPL estimators with different training data, from original training data of ${\textit{ControlEvents}}^{\text{SMPL}}$ to synthetic event data from baselines~\cite{zhu2021eventgan, xue2024elnr} and our generated data.
Quantitative results in~\cref{tab:compare_smpl_recovery} show that our generated data outperforms synthetic data from Event-GAN~\cite{zhu2021eventgan} and from SMPL-ESIM~\cite{xue2024elnr}. It also shows that our generated data can be combined with original training data to further enhance the 3D SMPL estimation performance.

\begin{figure}[t]
    \centering
    \captionsetup{type=figure}
    \includegraphics[width=\linewidth]{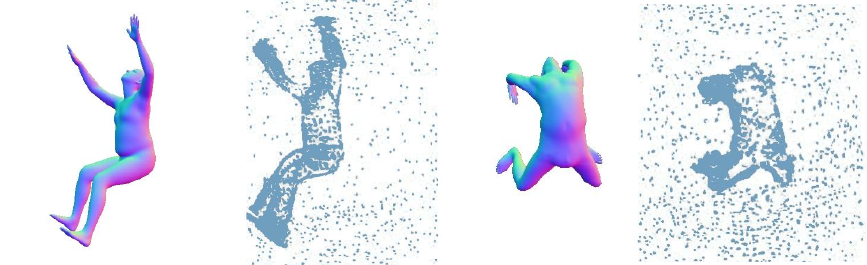}
    \vspace{-7mm}
    \caption{ \footnotesize \textbf{SMPL-based generation on challenging unseen poses}. Given challenging AMASS~\cite{mahmood2019amass} poses, we can generate realistic event data.}
    \label{fig:ood_poses}
    
    \vspace{-5mm}
\end{figure}

\vspace{-0.4cm}
\paragraph{Synthesizing Events for OOD Poses}
Generalizing to challenging out-of-distribution (OOD) poses remains a persistent issue in the pose recovery field, largely due to the scarcity of labeled training data for such complex poses, making this task inherently difficult.
As illustrated in~\cref{fig:ood_poses}, we demonstrate that our ${\textit{ControlEvents}}^{\text{SMPL}}$ generalizes effectively to challenging poses from the AMASS dataset~\cite{mahmood2019amass}. 
Moreover, we quantitatively validate in \cref{tab:compare_smpl_recovery} the benefits of training a SMPL estimator with our synthesized event data using pseudo ground-truth derived from AMASS.
Training solely with AMASS synthetic data yields slightly less accurate results, primarily due to differing pose distributions between the challenging AMASS dataset~\cite{mahmood2019amass} and the Event-HPE evaluation set~\cite{zou2021eventhpe}. However, training on a combination of real and challenging synthetic event data significantly improves SMPL recovery performance with a {$31.7\%$} reduction in 3D-MPJPE.

\vspace{-0.4cm}
\paragraph{Text-to-Events in Motion}
Our ${\textit{ControlEvents}}^{\text{SMPL}}$ can synthesize realistic event data from arbitrary SMPL annotations, providing the potential to leverage powerful text-to-motion models~\cite{tevet2022mdm, li2024unimotion} for generating human motion events based on text descriptions. In contrast, other learning-based methods, such as~\cite{zhu2021eventgan}, which generate events from intensity frames, cannot directly condition on SMPL for generation. Please refer to the supplementary video for animations.

\subsection{Event Generation Quality}
We quantitatively evaluate the quality of generated event data from our method compared to baselines under diverse conditions, as shown in~\cref{tab:FID_3Dsmpl}. 
As a reference, we provide the FID~\cite{heusel2017fid} scores of random unseen real event data relative to the training data distribution of generative models from the same datasets.
As shown in~\cref{fig:intro} and~\cref{tab:FID_3Dsmpl}, the event data generated by {\textit{ControlEvents}} is closest in distribution to the training data and closely approximates real captured event data, demonstrating the effectiveness of our method in generating high-quality synthetic events.

\begin{table}[h]
\resizebox{0.48\textwidth}{!}{
\begin{tabular}{ l c c c }
 \hline
Method  & $\text{FID}_{\text{SMPL}}\downarrow$  & $\text{FID}_{\text{skeleton}}\downarrow$ & $\text{FID}_{\text{class}}\downarrow$ \\
 \hline
Unseen real event data & $3.64$ & $23.22$ & $0.36$ \\
\hline
Event-GAN~\cite{zhu2021eventgan} &  $196.41$ & $165.93$  & $-$ \\
SMPL-ESIM~\cite{xue2024elnr}  & $206.23$  & $-$ & $-$  \\
\hline
\textit{Ours}  & $\textbf{15.55}$ & $\textbf{28.97}$ & $\textbf{27.43}$\\
 \hline
\end{tabular}}
\caption{\textbf{FID score between synthesized event and real-world event data.} We evaluate the quality of conditional distribution modelling w.r.t. SMPL, skeleton, and class label. Our synthetic event is more close to real event data compared to~\cite{zhu2021eventgan, xue2022events, xue2024elnr}.}
\label{tab:FID_3Dsmpl}
\vspace{-3mm}
\end{table}

\begin{figure}[h]
    \centering
    \captionsetup{type=figure}
    \includegraphics[width=\linewidth]{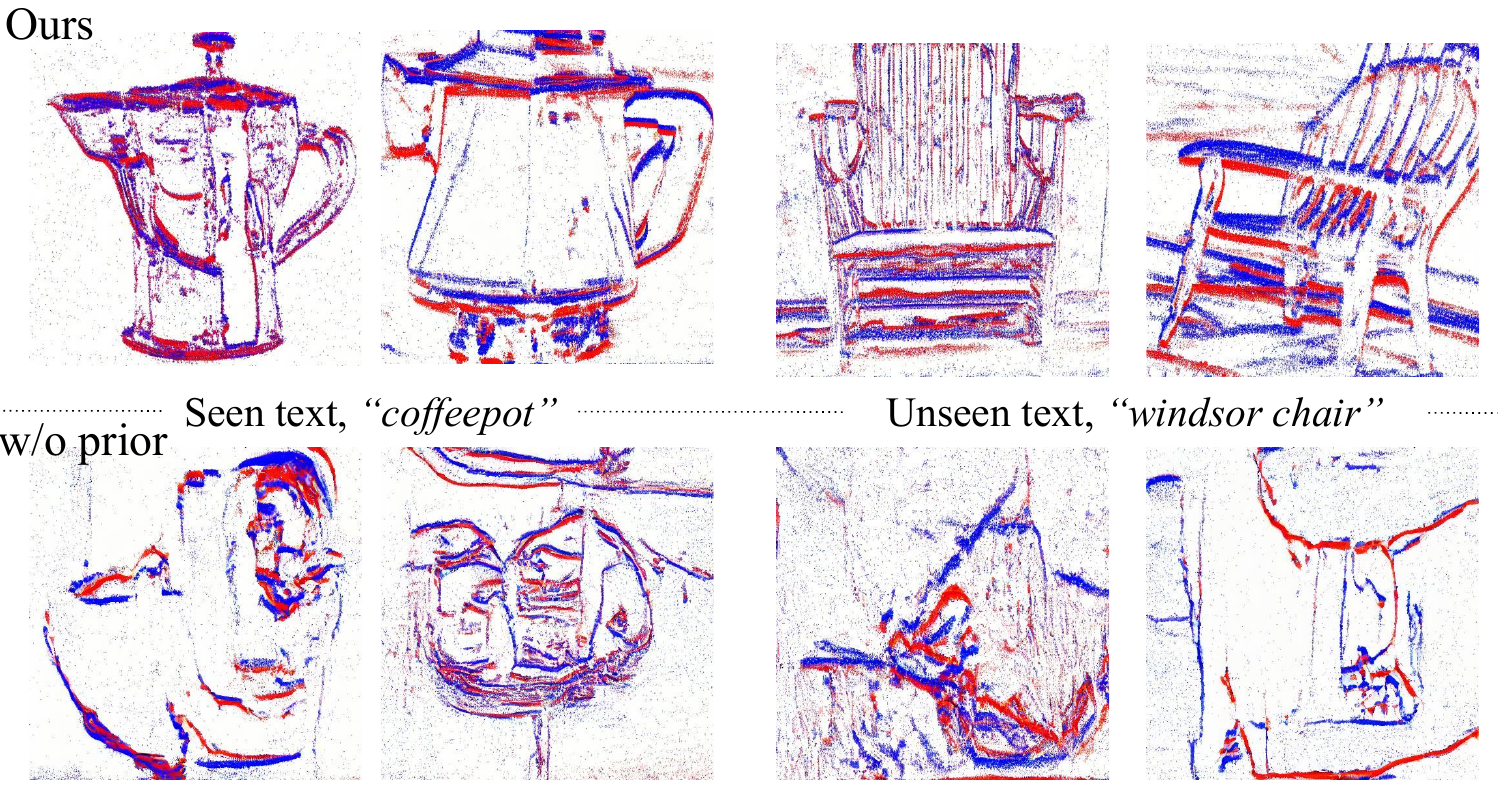}
    \vspace{-7mm}
    \caption{ \footnotesize  \textbf{Ablation on foundational prior from Stable Diffusion~\cite{rombach2021stablediffusion}}. Without the prior, our method cannot generate meaningful on unseen text.}
    \label{fig:ablation_random}
    \vspace{-6mm}
\end{figure}

\begin{figure*}[t]
    \centering
    \captionsetup{type=figure}
    \includegraphics[width=\linewidth]{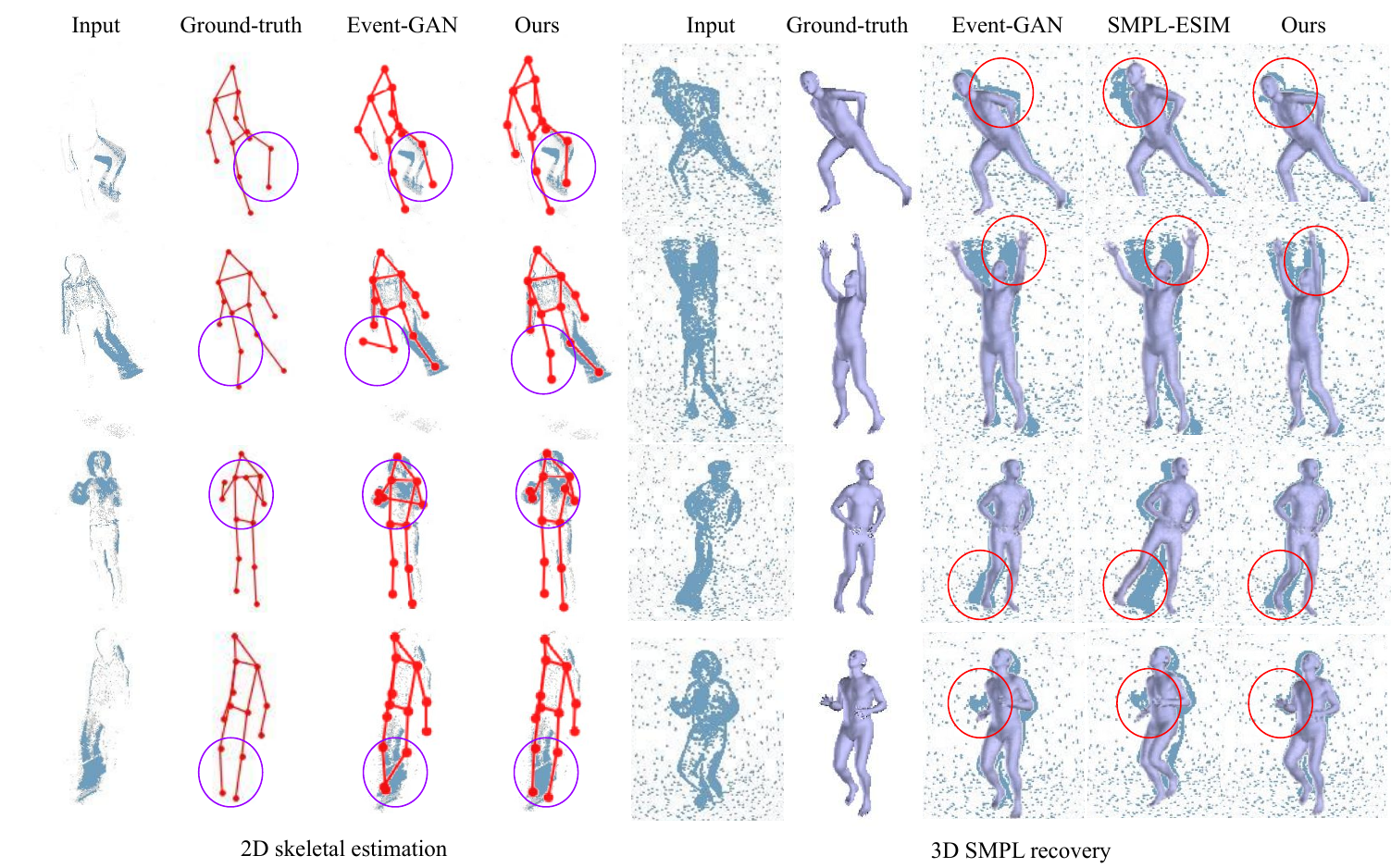}
    \vspace{-7mm}
    \caption{ \footnotesize Comparison synthetic data between \textbf{\textit{ControlEvents}} and other event simulators (Event-GAN~\cite{zhu2021eventgan}, SMPL-ESIM~\cite{xue2024elnr}) for 2D skeletal estimation and 3D body recovery. Estimators trained on \textit{ControlEvents} generated data shows better performance due to high quality data, especially on unseen parts.}
    \label{fig:comparison}
    \vspace{-5mm}
\end{figure*}

\subsection{Ablation Study}
\paragraph{Diffusion Prior from Foundation Models} A key insight in {\textit{ControlEvents}} is leveraging the powerful diffusion prior in pre-trained foundation models, enabling fine-tuning with minimal data while inheriting the capabilities of the original models. To demonstrate the effectiveness of this approach, we present \textit{Zero-Shot Generation} experiments in~\cref{subsec:exp_classification}, showing our model’s impressive text-to-image capabilities by generalizing to unseen text labels. Here, we further examine the benefits of the diffusion prior by conducting an ablation study on the diffusion prior from pre-trained foundation models.

Using the same training data $\mathcal{D}^{\text{class}}_{\text{real}}$ as experiments in~\cref{subsec:exp_classification}, we directly train a ${\textit{ControlEvents}}^{\text{class}}$ model without a foundational diffusion prior by randomly initializing the U-Net weights of Stable Diffusion~\cite{rombach2021stablediffusion}. In~\cref{fig:ablation_random}, we visualize the generation results at intermediate steps during training. The results intuitively show that, without the diffusion prior, the generative model struggles to learn the distribution of the training event data. We also observe that the model fails to generate meaningful event images for unseen text labels, producing only noise instead.

\begin{figure}[h]
    \centering
    \captionsetup{type=figure}
    \includegraphics[width=0.9\linewidth]{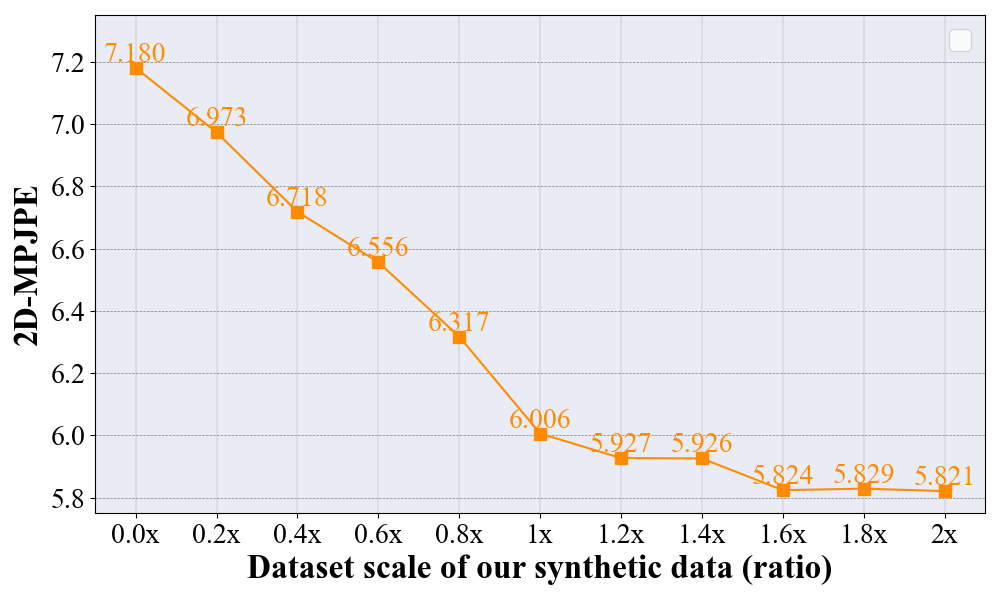}
    \vspace{-3mm}
    \caption{ \footnotesize \textbf{Ablation on scaling synthetic data of skeleton estimation}. Our synthesized events with pseudo ground-truth can enhance the skeleton estimation along with original training dataset.}
    \label{fig:scaling_2dskeleton}
    
    \vspace{-5mm}
\end{figure}

\vspace{-0.4cm}
\paragraph{Scaling of Synthesized Training Data} An advantage of {\textit{ControlEvents}} its ability to synthesize realistic event data with pseudo ground-truth labels at low cost. We conducted an ablation study to examine the impact of scaling the amount of synthesized event data from ${\textit{ControlEvents}}^{\text{skeleton}}$ for training a 2D pose estimator by scaling the dataset size from 0 (0.0x) to 20,000 (2.0x) samples, as illustrated in~\cref{fig:scaling_2dskeleton}. The results indicate that up-scaling the synthesized data can enhance the model performance.

\section{Conclusion}
\label{sec:conclusion}

We present \textbf{\textit{ControlEvents}}, a controllable event synthesis approach that leverages the foundational priors of pre-trained image diffusion models. We demonstrate that: 1) generating events with pseudo ground-truth labels can effectively enhance the performance of deep learning models, and 2) the diffusion prior from foundational models helps in generalizing to unseen scenarios for events generation tasks. We believe that our approach paves the way for new directions in data synthesis for event-based vision tasks.
We will make our model and synthetic event dataset for all tasks available for future research.

\vspace{-0.4cm}
\paragraph{Limitations and future work}
To best leverage the priors from image diffusion models, we accumulate event streams into images, which may result in the loss of rich temporal information inherent in asynchronous streams. A promising direction for future exploration is to utilize priors from video diffusion models~\cite{brooks2024sora, ma2024followpose, ma2025followcreation, ma2024followyouremoji, ma2025followyourclick, ma2025followyourmotion}, which can incorporate temporal information and potentially handle the time steps of event data more effectively.

{\footnotesize
\paragraph{Acknowledgements}
We thank G.Tiwari, I.Sarandi, V.Guzov, Y.He, X.Xie, C.Li whose feedback helped improve this paper. 
This work is made possible by funding from the Carl Zeiss Foundation. 
This work is also funded by the Deutsche Forschungsgemeinschaft (DFG, German Research Foundation) - 409792180 (EmmyNoether Programme, project: Real Virtual Humans) and the German Federal Ministry of Education and Research (BMBF): Tübingen AI Center, FKZ: 01IS18039A. 
The authors thank the International Max Planck Research School for Intelligent Systems (IMPRS-IS) for supporting Y.Xue.
G. Pons-Moll is a member of the Machine Learning Cluster of Excellence, EXC number 2064/1 – Project number 390727645. 
}

{\footnotesize
Y.Hu and Y.Xue contributed equally as the joint first authors. Y.Xue is the corresponding author. Authors with equal contributions are listed in alphabetical order and allowed to change their orders freely on their resume and website.  Y.Xue initialized the core idea, organized the project, co-developed the current method, co-supervised the experiments, and wrote the draft. Y.Hu co-developed the current method, implemented most of the prototypes, and conducted experiments. 
}

\begin{appendix}
\section{Dataset}
\label{supp:dataset}

\subsection{Dataset Description}

\paragraph{Classification}
In this paper, we utilize N-ImageNet~\cite{kim2021nimagenet} and N-Caltech101~\cite{orchard2015ncaltech} as the event data classification datasets. Both datasets are event-based conversions of ImageNet~\cite{kim2021nimagenet} and Caltech101~\cite{li2004caltech101}, created by capturing original RGB images displayed on a monitor using an event camera.

\begin{figure}[h]
    \centering
    \captionsetup{type=figure}
    \includegraphics[width=\linewidth]{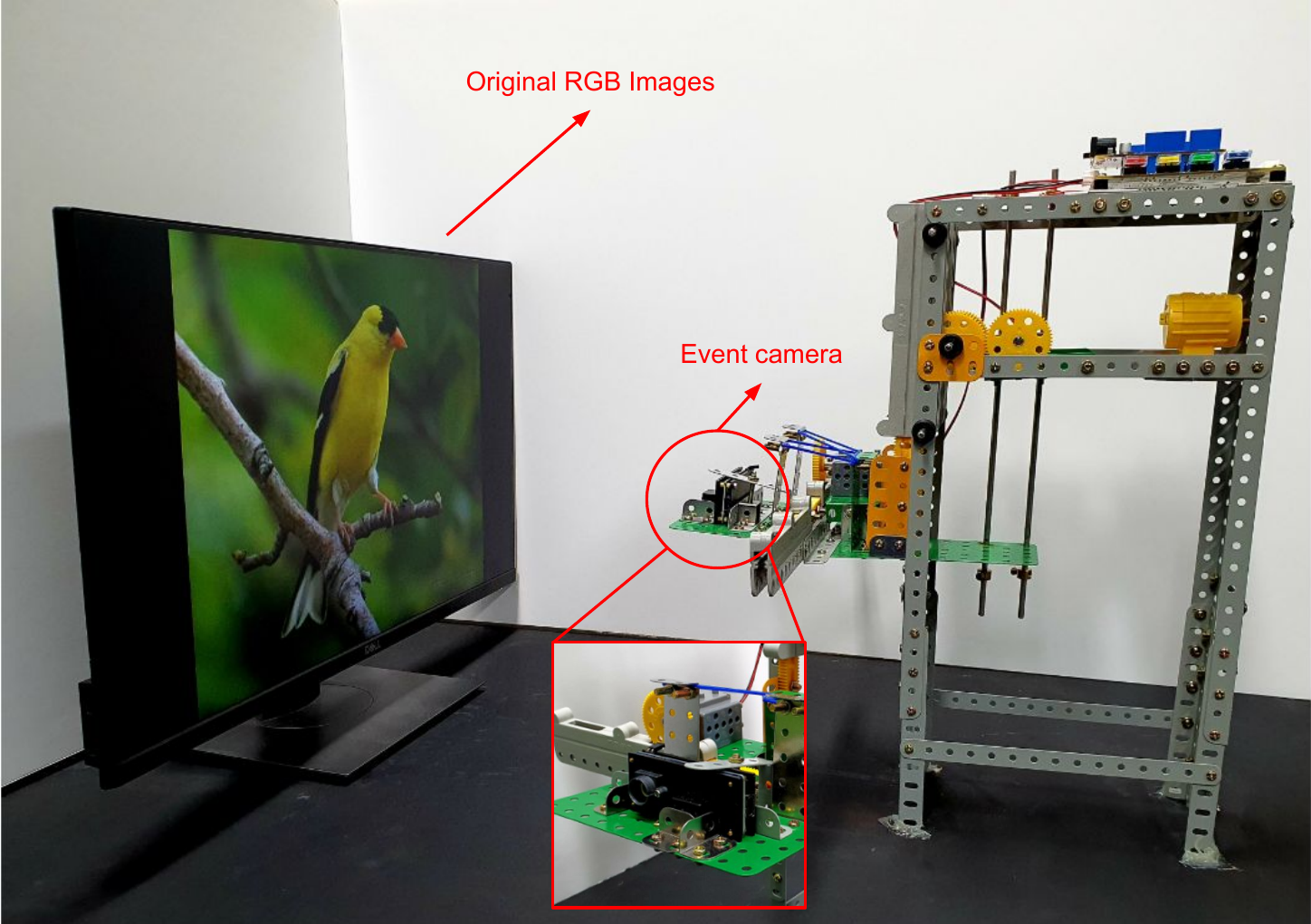}
    
    \caption{ \footnotesize \textbf{Conversion hardware system} in N-ImageNet~\cite{kim2021nimagenet}. The event camera is vibrated vertically and horizontally on a plane parallel to the LCD monitor screen to capture events of displayed images.}
    \label{fig:nimagenet_system}
    \vspace{-5mm}
\end{figure}

These datasets generate events from static RGB images by introducing relative camera-to-image motion, achieved either by moving the event cameras~\cite{kim2021nimagenet} or by moving the monitor~\cite{orchard2015ncaltech}. Details of the event acquisition system used in N-ImageNet~\cite{kim2021nimagenet} are illustrated in~\cref{fig:nimagenet_system}.

\vspace{1mm}
\subsection{Processing Datasets}

\begin{figure*}[t]
    \centering
    \captionsetup{type=figure}
    \includegraphics[width=\linewidth]{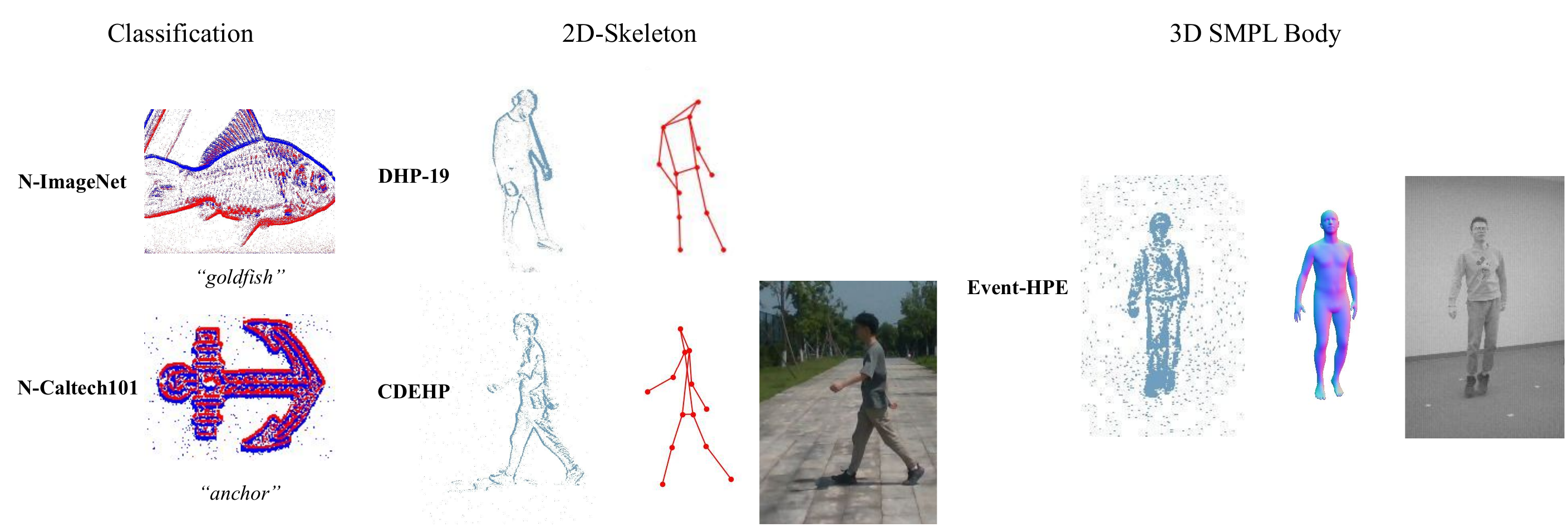}
     \vspace{-5mm}
    \caption{ \footnotesize \textbf{Visualization of processed datasets.} N-ImageNet~\cite{kim2021nimagenet} and N-Caltech101~\cite{orchard2015ncaltech} provide polarities of events. CDEHP~\cite{shao2024cdehp} and Event-HPE~\cite{zou2021eventhpe} provide intensity frames.}
    \label{fig:dataset}
    \vspace{-5mm}
\end{figure*}

As described in Section 3.1 of the main paper, we convert asynchronous events into RGB images to leverage the priors from pre-trained foundation models. We visualize the processed datasets in~\cref{fig:dataset}, including N-ImageNet~\cite{kim2021nimagenet}, N-Caltech101~\cite{orchard2015ncaltech}, DHP-19~\cite{calabrese2019dhp19}, CDEHP~\cite{shao2024cdehp}, and Event-HPE~\cite{zou2021eventhpe}.
The processing procedure is described as:
\begin{itemize} 
\item N-ImageNet: All events are accumulated into a single image, capturing the original ImageNet image. Positive events are represented in red, and negative events in blue. 
\item N-Caltech101: Similar to N-ImageNet, all events are accumulated into a single image. Positive events are represented in red, and negative events in blue. 
\item DHP-19: Following the original paper~\cite{calabrese2019dhp19}, 7,500 events are accumulated to construct each event image. We follow the original paper and discard polarities. DHP-19 includes 2D skeleton annotations as ground truth but does not include intensity frames. 
\item CDEHP: Following the original paper~\cite{shao2024cdehp}, events are accumulated over 8.333 ms to construct each event image. We follow the  CDEHP includes both 2D skeleton annotations and intensity frames. 
\item Event-HPE: Following the original paper~\cite{zou2021eventhpe}, events are accumulated over 15 ms to construct each event image. We follow the original paper and discard polarities of events. The dataset provides both 3D SMPL annotations and intensity frames. 
\end{itemize}

\section{Method}
\label{supp:method}

\subsection{Training details}

\paragraph{${\textit{ControlEvents}}^{\text{class}}$}
We train our text-based event generative model by adopting the Stable Diffusion v1.4 text-to-image checkpoint~\cite{stablediffusionv1} and fine-tuning it on our collected event data. The model is trained on 4 NVIDIA A40 GPUs for 80 epochs, with a batch size of 4 and gradient accumulation over 32 steps, resulting in an effective batch size of 512. The training process takes approximately 5 days.

\paragraph{${\textit{ControlEvents}}^{\text{skeleton}}$}
We train our skeleton-based event generative model by adopting the ControlNet human skeleton v1.0 checkpoint~\cite{controlnetskeleton} and fine-tuning it on our collected event data. The model is trained on 4 NVIDIA A40 GPUs for 600 epochs, with a batch size of 2 and gradient accumulation over 64 steps, resulting in an effective batch size of 512. The training process takes approximately 3 days.

\paragraph{${\textit{ControlEvents}}^{\text{SMPL}}$}

We train our SMPL-based event generative model by adopting the ControlNet normal v1.0 checkpoint~\cite{controlnetnormal} and fine-tuning it on our collected event data, using normal maps rendered from SMPL annotations as input. The model is trained on 4 NVIDIA A40 GPUs for 600 epochs, with a batch size of 2 and gradient accumulation over 64 steps, resulting in an effective batch size of 512. The entire training process takes approximately 3 days.

\vspace{2mm}
\subsection{Inference details}

At inference time, we use DDPM~\cite{Ho2020DDPM} as scheduler and set reverse sampling steps to 1000. The whole inference time on NVIDIA A40 (48GB) for ${\textit{ControlEvents}}^{\text{class}}$ takes 12 seconds,  for ${\textit{ControlEvents}}^{\text{skeleton}}$ takes 17s, and for ${\textit{ControlEvents}}^{\text{SMPL}}$ takes 17s. 

\section{Experiments}
\label{sup:gen}

\subsection{Generation from Text}
\begin{figure*}[t]
    \centering
    \captionsetup{type=figure}
    \includegraphics[width=\linewidth]{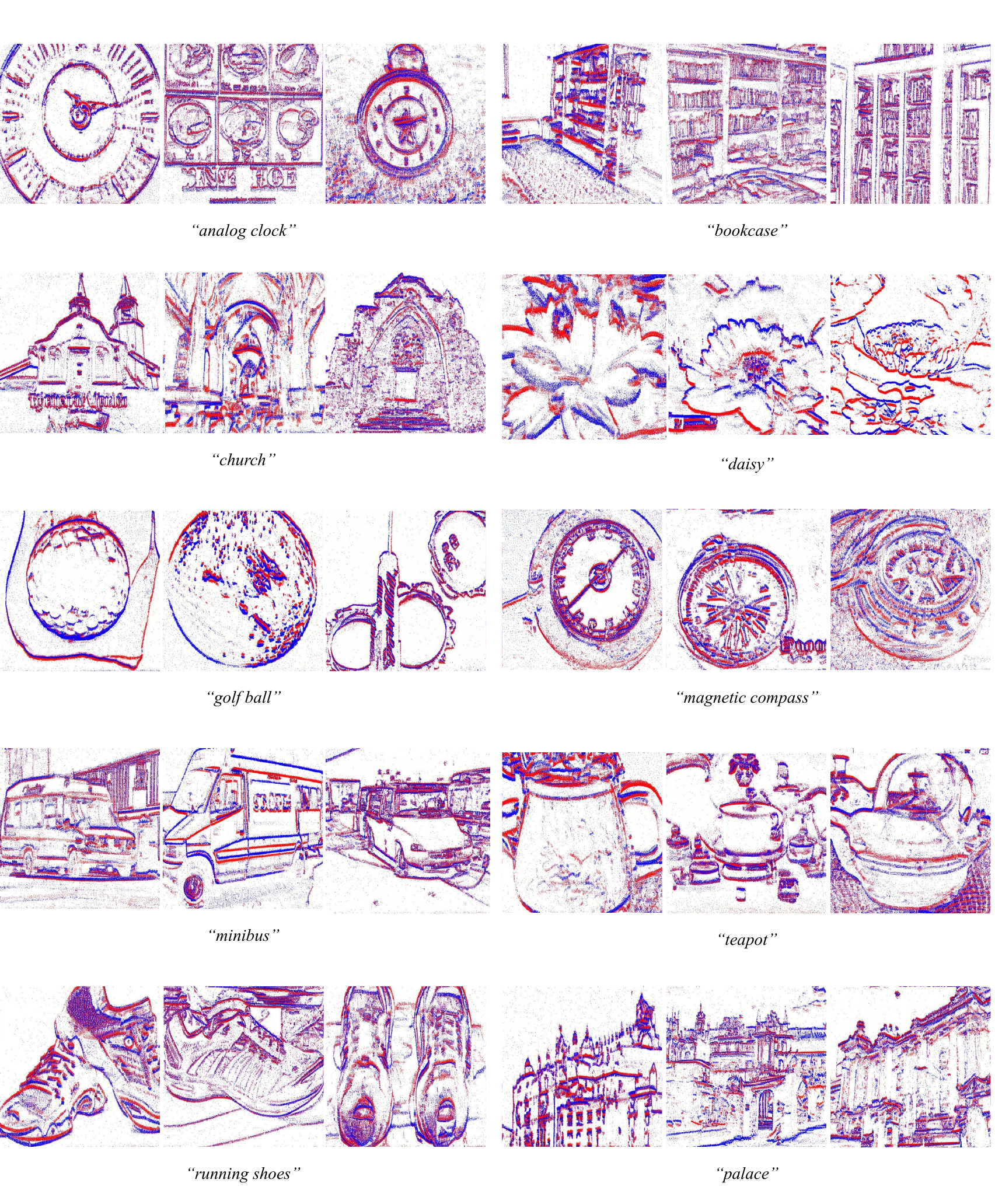}
    \caption{ \footnotesize \textbf{Generation of event images from text class labels} in N-ImageNet~\cite{kim2021nimagenet}.}
    \label{fig:text_conditioned}
\end{figure*}

In the main paper, we only present the generated event images from text labels in fig.1. Thus, we visualize more text-conditioned generation results in~\cref{fig:text_conditioned}. Please refer to our supplementary video for more visualization results.

\subsection{Zero-Shot Text-based Generation}
\begin{figure*}[t]
    \centering
    \captionsetup{type=figure}
    \includegraphics[width=\linewidth]{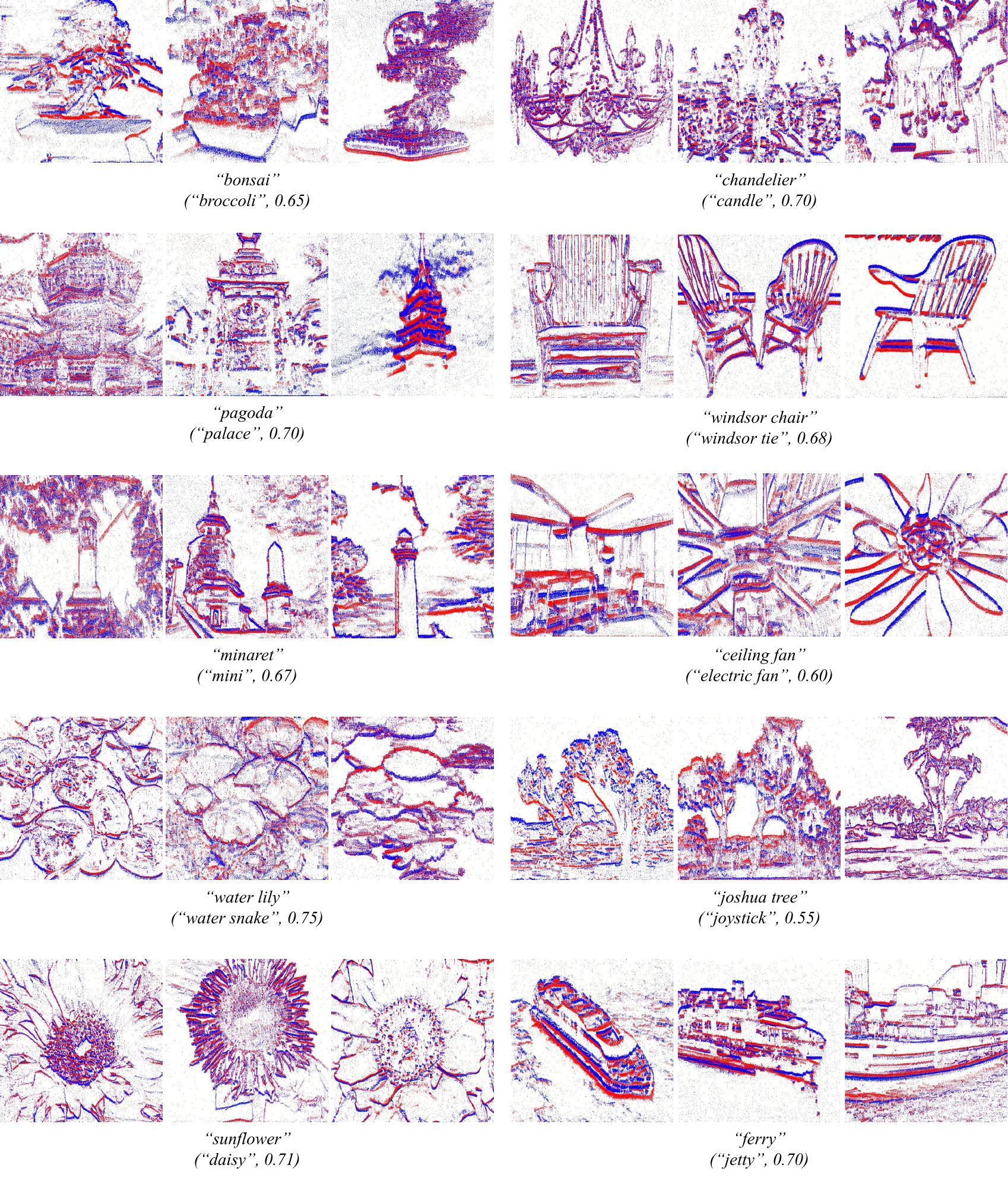}
    \caption{ \footnotesize \textbf{Zero-shot generation of unseen text label} from N-Caltech101~\cite{orchard2015ncaltech} dataset. We determine the closest seen text label during training based on the CLIP cosine similarity.}
    \label{fig:zeroshot}
\end{figure*}

As introduced in section 4.1, our ${\textit{ControlEvents}}^{\text{class}}$ can generate event images from text labels that were unseen during training on the N-ImageNet~\cite{kim2021nimagenet} dataset. Here, we present additional qualitative zero-shot binary classification results (see ~\cref{fig:zeroshot}) for 10 unseen classes from the N-Caltech101~\cite{orchard2015ncaltech} dataset. In~\cref{fig:zeroshot} we show the closest seen text label from N-ImageNet~\cite{kim2021nimagenet} and their cosine similarity in the CLIP~\cite{radford2021clip} space.
We use ${\textit{ControlEvents}}^{\text{class}}$ to generate 500 event images for each class. For comparison, we train the baseline classification model using 50 event images per class from the N-Caltech101 dataset. Quantitative results demonstrate that our zero-shot generated large-scale dataset outperforms the few-shot classification baseline trained on limited data.

\subsection{Generation from Skeleton \& SMPL}
\begin{figure*}[t]
    \centering
    \captionsetup{type=figure}
    \includegraphics[width=\linewidth]{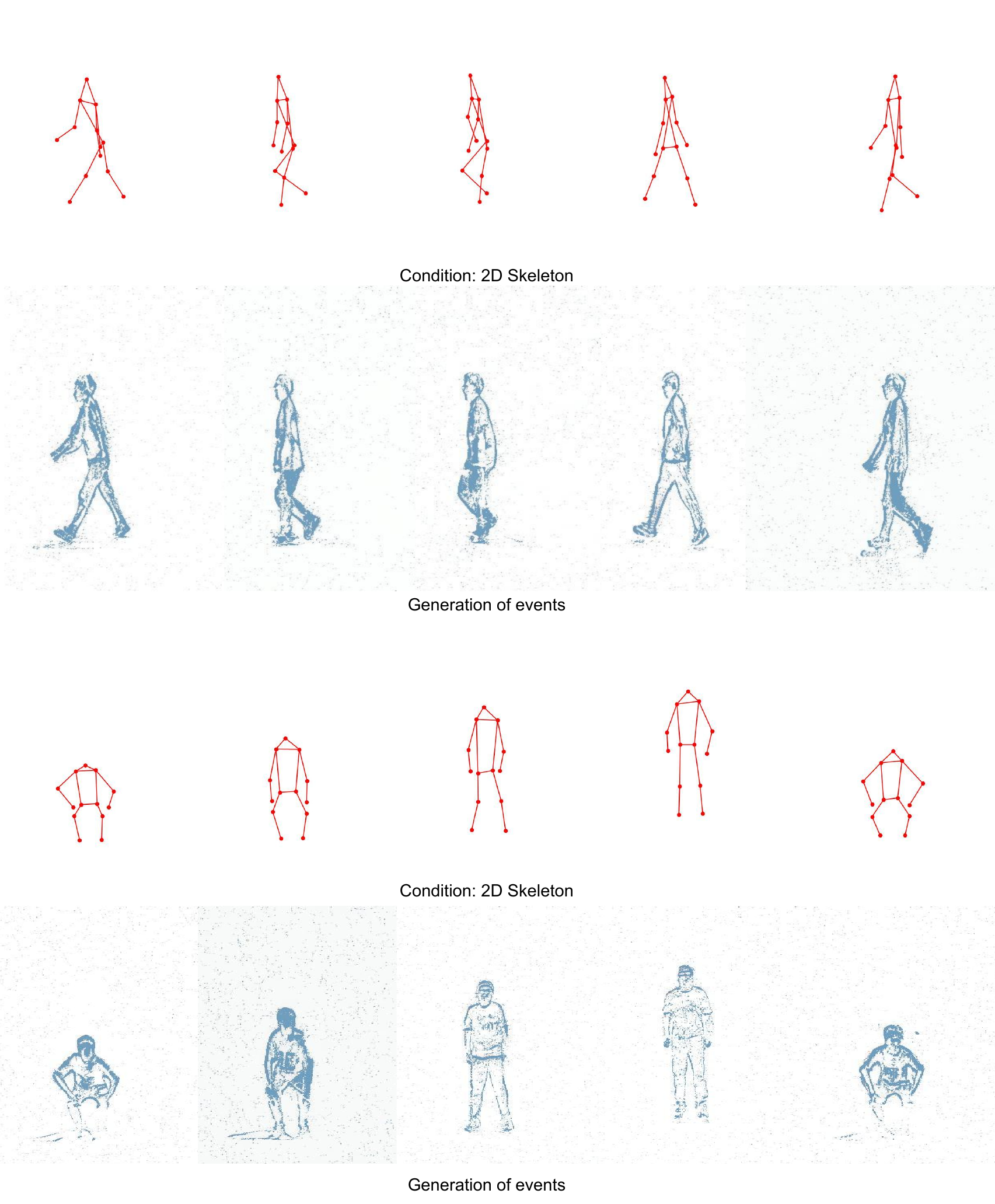}
    \caption{ \footnotesize \textbf{Generation of event images from 2D skeleton.} Please refer to supplementary video for animation results.}
    \label{fig:skeleton_conditioned}
\end{figure*}

\begin{figure*}[t]
    \centering
    \captionsetup{type=figure}
    \includegraphics[width=\linewidth]{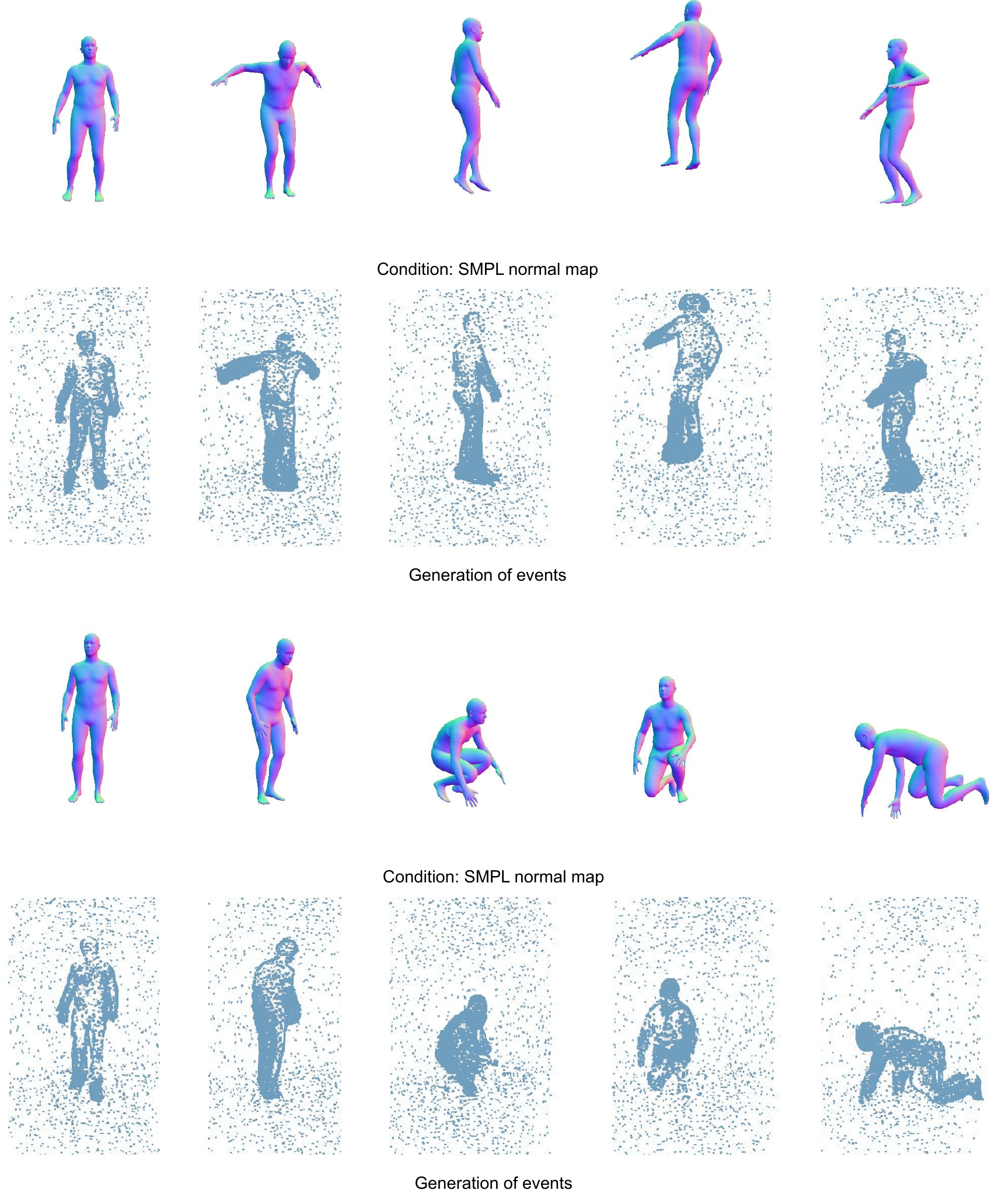}
    \caption{ \footnotesize \textbf{Generation of event images from 3D SMPL normal maps.} Please refer to supplementary video for animation results.}
    \label{fig:smpl_conditioned}
\end{figure*}

We visualize more skeleton-conditioned event images generation results in~\cref{fig:skeleton_conditioned} and SMPL-conditioned generation results in~\cref{fig:smpl_conditioned}. Please refer to our supplementary video for animation results.
We also present the animation results of text-to-events in motion, which we describe in Sec.4.3 in the main paper. Please refer to our supplementary video for the details.

\vspace{-2mm}

\begin{table}[h]
  \centering
{
\begin{tabular}{l c c}
 \hline
\multicolumn{3}{c}{N-Caltech101 zero-shot set~\cite{orchard2015ncaltech}} \\
 \hline
Class & Baseline {Acc. $\uparrow$}  & Our {Acc. $\uparrow$} \\
 \hline
\textit{"bonsai"}   &    $ 55.00\% $   & $\textbf{76.25\%}$  \\
\textit{"ceiling fan"}   &    $ \textbf{85.06\%} $   & $ 67.82\% $  \\
\textit{"ferry"}   &   $  60.22\% $   & $ \textbf{68.82\%} $  \\
\textit{"joshua tree"}  &   $ 61.54\% $   & $ \textbf{70.51\%} $  \\
\textit{"minaret"}  &   $ 54.44\% $   & $ \textbf{60.00\%} $  \\
\textit{"pagoda"}   &   $ 65.15\% $   & $ \textbf{80.30\%} $  \\
\textit{"sunflower"}  &   $ 60.00\% $   & $ \textbf{74.67\%} $  \\
\textit{"water lilly"}  &   $ \textbf{72.41\%} $   & $ 55.17\% $  \\
\textit{"windsor chair"}   &  $ \textbf{62.16\%} $   & $ 58.11\% $  \\
\textit{"chandelier"}  &  $ 58.02\% $   & $ \textbf{70.37\%} $  \\
\hline
\end{tabular}
}

\caption{\textbf{Zero-shot binary classification} on unseen classes from N-Caltech101~\cite{orchard2015ncaltech}.}
\label{tab:zeroshot_classification}
\vspace{-3mm}
\end{table}

\end{appendix}

{
    \small
    \bibliographystyle{ieeenat_fullname}
    \bibliography{main}
}

\end{document}